\documentclass[10pt,journal]{IEEEtran}
\usepackage{amsmath,amsfonts}
\usepackage{algorithmic}
\usepackage{algorithm}
\usepackage{array}
\ifCLASSOPTIONcompsoc
\usepackage[caption=false, font=normalsize, labelfont=sf, textfont=sf]{subfig}
\else
\usepackage[caption=false, font=footnotesize]{subfig}
\fi
\usepackage{textcomp}
\usepackage{stfloats}
\usepackage{url}
\usepackage{booktabs} 
\usepackage{verbatim}
\usepackage{graphicx}
\usepackage{framed}
\usepackage{color}
\usepackage{cite}

\usepackage{soul}

\definecolor{byzantine}{rgb}{0.74, 0.2, 0.64}

\graphicspath{{Figures/}}
\usepackage{hyperref}

\begin{document}

\title{Benchmarking Ultra-High-Definition Image Reflection Removal}

\author{Zhenyuan Zhang, Zhenbo Song, Kaihao Zhang, Zhaoxin Fan, Jianfeng Lu
\thanks{Z. Zhang, Z. Song and J. Lu are with the School of Computer Science and Engineering, Nanjing University of Science and Technology, Nanjing 210094, China. (Email: zyZhang.bbetter@gmail.com; songzb@njust.edu.cn; lujf@njust.edu.cn)}
\thanks{Kaihao Zhang is with the College of Engineering and Computer Science, Australian National University, Canberra, ACT, Australia.  (Email: super.khzhang@gmail.com)}
\thanks{Zhaoxin Fan is with Renmin University of China, Beijing 100872, China
(Email: fanzhaoxin@ruc.edu.cn)}
}
\markboth{Journal of \LaTeX\ Class Files,~Vol.~14, No.~8, August~2021}%
{Shell \MakeLowercase{\textit{et al.}}: A Sample Article Using IEEEtran.cls for IEEE Journals}

\IEEEpubid{0000--0000/00\$00.00~\copyright~2021 IEEE}


\maketitle

\begin{abstract}
	%
	Deep learning based methods have achieved significant success in the task of single image reflection removal (SIRR). However, the majority of these methods are focused on High-Definition/Standard-Definition (HD/SD) images, while ignoring higher resolution images such as Ultra-High-Definition (UHD) images. With the increasing prevalence of UHD images captured by modern devices, in this paper, we aim to address the problem of UHD SIRR. Specifically, we first synthesize two large-scale UHD datasets, UHDRR4K and UHDRR8K. The UHDRR4K dataset consists of $2,999$ and $168$ quadruplets of images for training and testing respectively, and the UHDRR8K dataset contains $1,014$ and $105$ quadruplets. To the best of our knowledge, these two datasets are the first largest-scale UHD datasets for SIRR. Then, we conduct a comprehensive evaluation of six state-of-the-art SIRR methods using the proposed datasets. Based on the results, we provide detailed discussions regarding the strengths and limitations of these methods when applied to UHD images. Finally, we present a transformer-based architecture named RRFormer for reflection removal. RRFormer comprises three modules, namely the Prepossessing Embedding Module, Self-attention Feature Extraction Module, and Multi-scale Spatial Feature Extraction Module. These modules extract hypercolumn features, global and partial attention features, and multi-scale spatial features, respectively. To ensure effective training, we utilize three terms in our loss function: pixel loss, feature loss, and adversarial loss. We demonstrate through experimental results that RRFormer achieves state-of-the-art performance on both the non-UHD dataset and our proposed UHDRR datasets. The code and datasets are publicly available at \href{https://github.com/Liar-zzy/Benchmarking-Ultra-High-Definition-Single-Image-Reflection-Removal}{https://github.com/Liar-zzy/Benchmarking-Ultra-High-Definition-Single-Image-Reflection-Removal}.
\end{abstract}


\section{Introduction}
\IEEEPARstart{T}{he} task of single image reflection removal (SIRR) is to recover a clear transmission image by removing reflection from the blended image. This task is of significant importance in computational photography, as it not only enhances image quality but also has a positive impact on downstream computer vision tasks, such as object detection \cite{chenHighQualityRCNNObject2021,9934934,9968016} and semantic segmentation \cite{wengStageAwareFeatureAlignment2022,maPRSegLightweightPatch2023}. 
Since the reflection removal problem is ill-posed, early works mainly focus on multi-image methods \cite{gaiBlindSeparationSuperimposed2011,guoRobustSeparationReflection2014,hanReflectionRemovalUsing2017,szeliskiLayerExtractionMultiple2000,sarelSeparatingTransparentLayers2004,liExploitingReflectionChange2013,xueComputationalApproachObstructionfree2015,sunAutomaticReflectionRemoval2016}. 
Recently, deep learning has been increasingly utilized for reflection removal, obviating the need for designing diverse priors.  With an adequate training dataset, deep learning models have demonstrated impressive outcomes \cite{fanGenericDeepArchitecture2017,wanCrrnMultiscaleGuided2018,johnsonPerceptualLossesRealtime2016,yangSeeingDeeplyBidirectionally2018,chiSingleImageReflection2018,jinLearningSeeReflections2018,leeGenerativeSingleImage2018,zhangSingleImageReflection2018,wanCoRRNCooperativeReflection2019,wenSingleImageReflection2019,weiSingleImageReflection2019,kimSingleImageReflection2020,liSingleImageReflection2020,leiRobustReflectionRemoval2021,liImageReflectionRemoval2022,songMultistageCurvatureGuidedNetwork2022,zhangSingleImageReflection2023,songRealSceneReflectionRemoval2023,fengU2FormerNestedUshaped2023,song2023robust}.

\begin{figure*}[!t]
	\centering
	\subfloat[Samples from the UHDRR4K dataset.]
	{
		\includegraphics[width=7in]{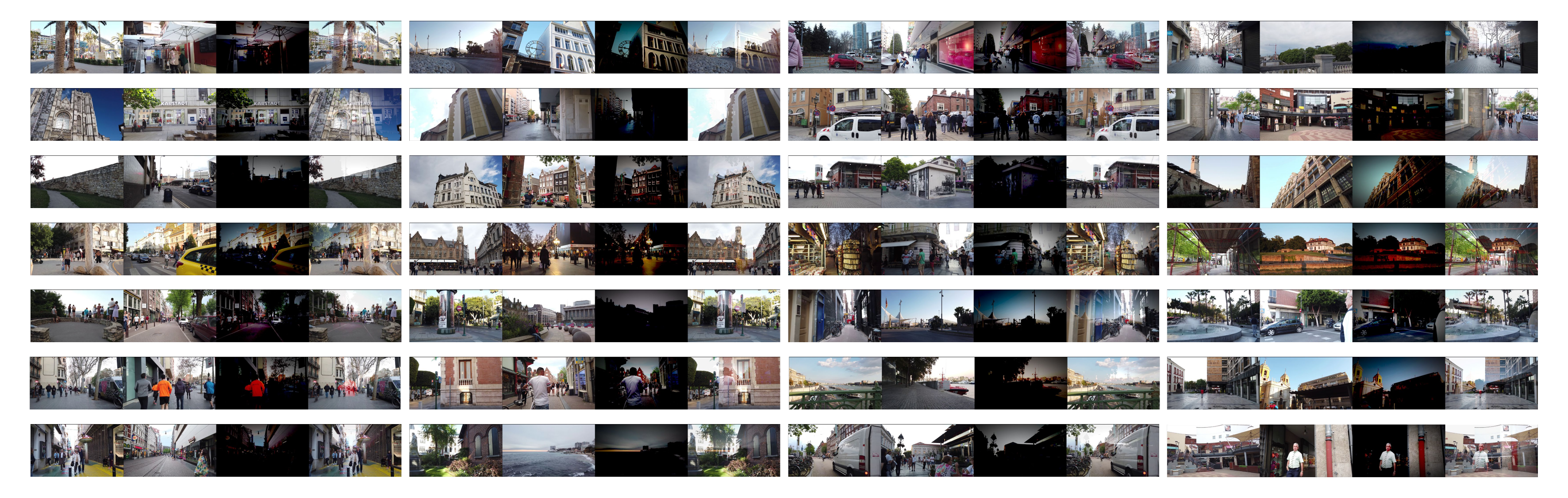}
		\label{fig_dataset_4k}
	}
	\hfil
	\subfloat[Samples from the UHDRR8K dataset.]
	{
		\includegraphics[width=7in]{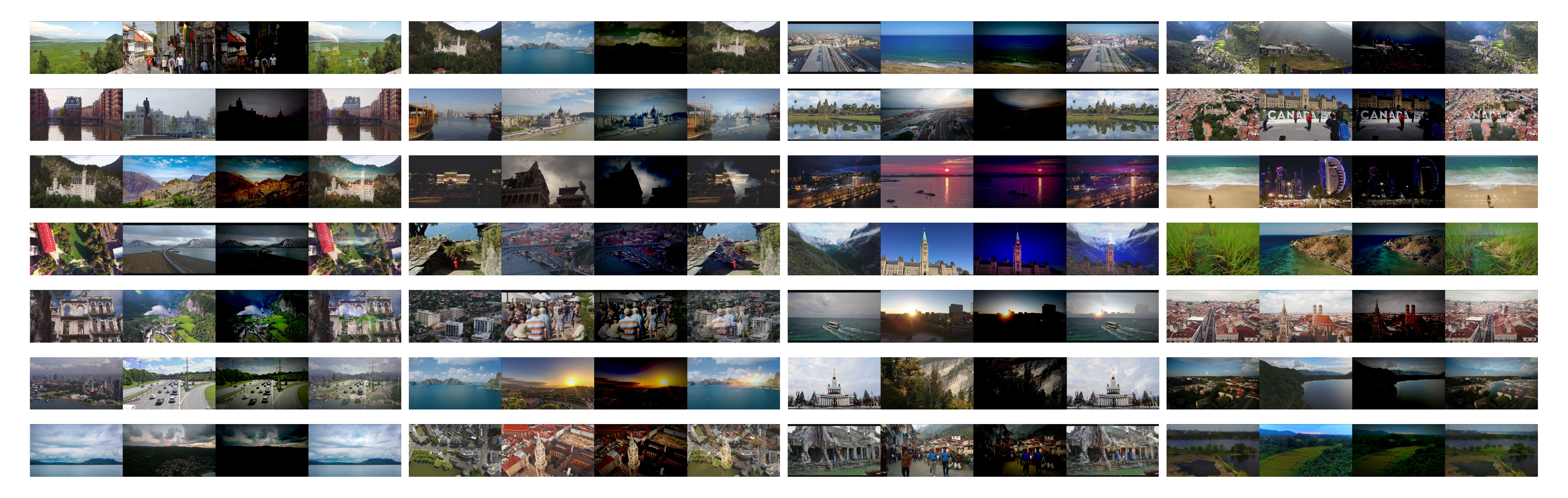}
		\label{fig_dataset_8k}
	}
	\caption{Sample images from the UHDRR4K and UHDRR8K datasets. These two datasets consist of a large number of 4K and 8K UHD images, respectively. Each sample is a quadruplet consisting of four images, \textit{i.e.}, transmission layer, reflection layer, reflection mask layer, and blended layer.}
	\label{fig_dataset}
\end{figure*}

Among these methods, most of them are trained and evaluated on natural images or synthetic images of SD or HD resolution. Hence, it is not clear how these methods perform on UHD images, \textit{e.g.}, 4K and 8K images. 
The majority of these methods are trained and evaluated on either natural or synthetic images of SD or HD resolution. Therefore, their performance on UHD images, such as 4K and 8K images, remains unclear. 
Considering that increasing mobile devices support capturing images of UHD resolution, 
this paper aims to study the problem of UHD SIRR. To investigate the performance of deep SIRR methods in UHD images, this paper first synthesizes two large-scale datasets, called UHDRR4K and UHDRR8K. 
The 4K dataset, \textsl{UHDRR4K}, includes $2,999$ and $168$ images for training and testing, respectively. The 8K dataset, \textsl{UHDRR8K}, contains $1,014$ training and $105$ testing images, respectively. 
To the best of our knowledge, UHDRR4K and UHDRR8K are the first large-scale datasets for UHD SIRR. 
Figure \ref{fig_dataset} provides samples from the constructed UHDRR4K and UHDRR8K datasets. Each training/testing sample is a quadruplet consisting of four images, \textit{i.e.}, ${T, R^*, R, B}$, where they respectively represent the transmission layer, reflection layer, reflection mask layer, and blended layer.

\IEEEpubidadjcol

\begin{figure}[t]
	\centering
	\includegraphics[width=3.5in]{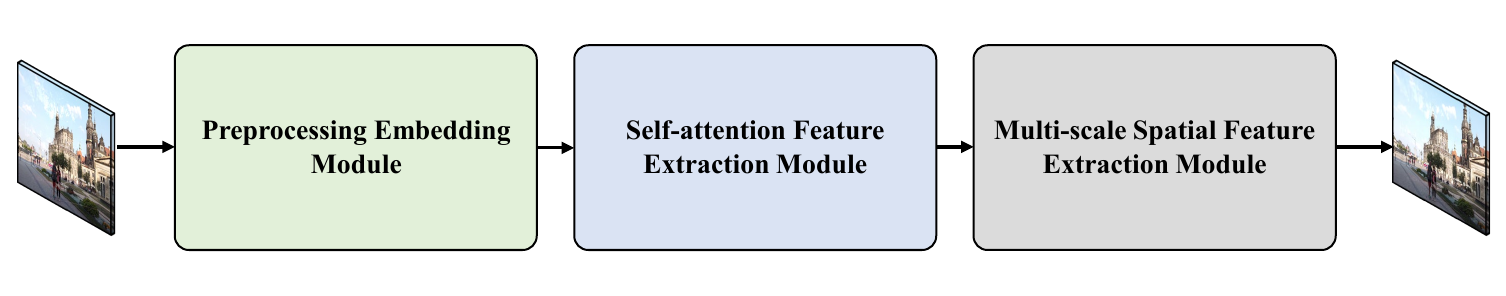}
	\caption{The proposed network RRFormer. RRFormer takes a single image as input, and remove the reflection contained in the given image through three modules, \textit{i.e.}, Preprocessing Embedding Module, Self-attention Feature Extraction Module and Multi-scale Spatial Feature Extraction Module.}
	\label{fig_network_overview}
\end{figure}

To explore the performance of current SIRR methods on UHD images, we evaluate six state-of-the-art methods on the two synthesize datasets. Standard metrics including PSNR (Peak Signal-to-Noise Ratio), SSIM (Structural Similarity Index Measure) \cite{wangImageQualityAssessment2004} and perceptual quality are used to evaluate how the state-of-the-art methods perform on UHD images.


Furthermore, as shown in Figure \ref{fig_network_overview}, we propose a transformer-based architecture named RRFormer, designed for SIRR. The RRFormer comprises three parts: Preprocessing Embedding Module, Feature Extraction Module, and Multi-scale Spatial Feature Extraction Module. In the Preprocessing Embedding Module, we leverage a pretrained VGG-19 network to extract hypercolumn features, which are then concatenated with the input blended image to create an enhanced input for the network. 
As for the Feature Extraction Module, we employ the residual swin transformer block \cite{liangSwinirImageRestoration2021} to extract global features. 
Finally, a pyramid pooling module \cite{heSpatialPyramidPooling2015,zhaoPyramidSceneParsing2017} is applied in the Multi-scale Spatial Feature Extraction Module to aggregate contextual information. 
Compared with traditional CNN-based models, our RRFormer integrates an attention mechanism to enhance feature representation by capturing global pixel interactions. Experimental evaluations conduct on both existing and newly proposed UHD datasets demonstrate that our RRFormer surpasses existing methods, establishing its superiority in this field.

In summary, the major contributions of our work are summarized as follows.

\begin{itemize}
	\item \textbf{Two large-scale UHDRR datasets.} We synthesize two large-scale UHD image datasets for SIRR. To the best of our knowledge, they are the largest-scale UHD datasets for reflection removal in the community. Each of quadruplets contains of four images, \textit{i.e.}, transmission layer, reflection layer, reflection mask layer, and blended layer. 
	
	\item \textbf{Comprehensive quantitative and qualitative benchmarking studies.} We comprehensively investigate the performance of the state-of-the-art single image reflection removal methods on the two UHDRR datasets. The study results reveal the limitation of current methods and inspire future research. 
	
	\item \textbf{RRFormer.} A transformer-based architecture, namely RRFormer, is proposed for single image reflection removal. The extensive experiments on the non-UHD dataset, \textit{i.e.}, CDR and our proposed UHDRR datasets indicate that RRFormer achieves superior quantitative performance as well as higher perceptual quality. 
\end{itemize}

\section{Related Work}
\subsection{SIRR Datasets}
Several datasets are built for SIRR training and evaluation, including SIR$^2$ \cite{wanBenchmarkingSingleimageReflection2017}, Zhang et al. \cite{zhangSingleImageReflection2018}, Nature \cite{liSingleImageReflection2020}, CDR \cite{leiCategorizedReflectionRemoval2022}. 
Wan \textit{et al.} \cite{wanBenchmarkingSingleimageReflection2017} propose a dataset named SIR$^2$ for SIRR. 
This dataset contains $40$ controlled indoor scenes and $100$ wild scenes, each of which is a triplet including mixture image, transmission and reflection. When capturing indoor scenes, they control the scene with a set of solid objects, five postcards and their combinations. They also utilize different configurations of aperture size and exposure time to ensure constant brightness, and three glasses of different thicknesses to explore the effect of thickness. In terms of wild scenes, they also take into account the reflectivity of objects, different illuminations, distances and scales. 
Zhang \textit{et al.} \cite{zhangSingleImageReflection2018} use pairs of images from Flickr to synthesize the dataset. They believe that the transmission layer and the reflection layer have different blurriness. Based on this, they apply a random Gaussian smoothing kernel to the reflection layer, making the synthesized image more realistic. They also capture $110$ real image pairs by placing a portable glass in front of the camera. Environments, lighting conditions, capture angles and apertures are all taken into account for unique variables.
Similar to \cite{zhangSingleImageReflection2018}, Nature \cite{liSingleImageReflection2020} includes $220$ real-world pairs. Additionally, they also pay attention to different thickness of the glasses. 
The CDR dataset \cite{leiCategorizedReflectionRemoval2022} includes $1,063$ triplets of $M, R, T$ in the wild, where $M$, $R$ and $T$ are the mixed image, the reflection image and the transmission image. They adopt the M-R pipeline \cite{leiPolarizedReflectionRemoval2020} to capture perfectly aligned images. 
To achieve this, the researchers capture the reflection image by positioning a black cloth behind the glass and capturing the mixed image with the glass present. To ensure dataset diversity, various glasses, objects, and lighting conditions are employed. To investigate the impact of different methods on images with varying levels of difficulty, the dataset is divided into several sub-datasets based on smoothness, relative intensity, and ghosting.

Among these datasets, the image resolutions of Zhang \textit{et al.} \cite{zhangSingleImageReflection2018} and CDR \cite{leiCategorizedReflectionRemoval2022} are relatively large which range from $1152 \times 930$ to $2109 \times 1396$, but fail to meet the UHD standard. SIR$^2$ \cite{wanBenchmarkingSingleimageReflection2017} is a large-scale dataset with $1,200$ images, but the typical image resolution is only $1720 \times 1234$. 
In this paper, we first synthesize two new large-scale datasets, and then benchmark deep learning based UHD SIRR methods on 4K and 8K images. 
Compared with prior SIRR datasets, our datasets exhibit significantly higher resolution, as shown in Table \ref{table_dataset}.

\begin{table}[!t]
	\centering
	\caption{Representative single image refection removal datasets. We introduce two new large-scale UHD (4K and 8K) reflection removal benchmark datasets.}
	\label{table_dataset}
	\begin{tabular}{l|rrr}
		\toprule
		Dateset& Size & Avg. Resolution & Format\\
		\midrule
		SIR$^2$ \cite{wanBenchmarkingSingleimageReflection2017}&1,500&$1726\times1234$&JPG\\
		Zhang \textit{et al.} \cite{zhangSingleImageReflection2018}&110&$1, 152\times930$&JPG\\
		Nature \cite{liSingleImageReflection2020}&220&$600\times400$&JPG\\
		CDR \cite{leiCategorizedReflectionRemoval2022}&1, 063&$2, 109\times1, 396$&PNG\\
		\midrule
		UHDRR4K&3, 167&$3, 840\times2, 160$&PNG\\
		UHDRR8K&1, 119&$7, 680\times4, 320$&PNG\\
		\bottomrule
	\end{tabular}
\end{table}

\subsection{Traditional SIRR Methods}
Single image reflection removal is a massively ill-posed problem. Previous methods \cite{levinLearningPerceiveTransparency2002,levinUserAssistedSeparation2007,liSingleImageLayer2014,shihReflectionRemovalUsing2015,wanDepthFieldGuided2016,arvanitopoulosSingleImageReflection2017,wanRegionawareReflectionRemoval2018} rely on priors or other information to handle specific scenarios. 
To discover minimum edges and corners for layer decomposition, the widely used prior, natural image gradient sparsity \cite{levinLearningPerceiveTransparency2002} is applied.
An optimization model is built for gradients and cornerness in natural scenes to generate better predictions. 
Gradient sparsity priors are also explored along with optimal and minimal user assistance to better guide uncertain separation problems \cite{levinUserAssistedSeparation2007}. The iterative reweighted least squares (IRLS) is applied to the optimization model. To obtain better performance, manual markers consisting of certain edges (or regions) are utilized as a prior. However, this method is labor-intensive and leans to result in mistakes. 
\cite{liSingleImageLayer2014} utilizes the different gradients between the transmission layer and the reflection layer, while it reveals limitations in the scene of specular highlighting. 
In \cite{shihReflectionRemovalUsing2015}, reflection is removed by using ghosting effects and the Gaussian Mixture Model (GMM) \cite{zoranLearningModelsNatural2011} is adopted to validate the model. 
\cite{arvanitopoulosSingleImageReflection2017} addresses the optimization problem via a Laplacian data fidelity term and an $l_{0}$ prior term to suppress reflection. 
In \cite{wanRegionawareReflectionRemoval2018}, a region-aware reflection removal approach which combines content and gradient priors to achieve content recovery as well as reflection separation is proposed.
However, these methods heavily rely on scene priors. Different imaging conditions and complex scene content in the real-world make their generalizability problematic.

\subsection{Deep Learning based SIRR Methods}\label{SIRR_Methods}
Recently, there has been a growing interest in applying deep learning to reflection removal and most of the current state-of-the-art SIRR methods \cite{fanGenericDeepArchitecture2017,zhangSingleImageReflection2018,yangSeeingDeeplyBidirectionally2018,weiSingleImageReflection2019,wenSingleImageReflection2019,liSingleImageReflection2020,kimSingleImageReflection2020,changSingleImageReflection2021} are based on deep learning. 
CEILNet \cite{fanGenericDeepArchitecture2017} is the first to solve the task of single image reflection removal using deep neural networks. 
They utilize a deep network to predict the edge of the map, and then exploit predicted edge maps to predict the transmission layer. 
Later, the conditional GAN \cite{isolaImagetoimageTranslationConditional2017} is introduced by Zhang \textit{et al.} \cite{zhangSingleImageReflection2018} to predict the transmission layer realistically. They also propose feature loss and exclusion loss to better separate reflection layer and transmission layer.
BDN \cite{yangSeeingDeeplyBidirectionally2018} utilizes a cascade deep neural network to predict the reflection layer, which is then used as feature information to predict the transmission layer. 
Wei \textit{et al.} \cite{weiSingleImageReflection2019} propose a contextual sensitively network, which includes two contextual forms, channel-wise context and multi-scale spatial context. They also introduce an alignment-invariant loss for training misaligned data. 
Wen \textit{et al.} \cite{wenSingleImageReflection2019} present a synthesis network to predict a non-linear alpha blending mask and a removal network to predict the transmission layer which utilizes the predicted mask.
A cascaded network is proposed in IBCLN \cite{liSingleImageReflection2020} where an LSTM mutually improves the quality of the predicted transmission and reflection. Moreover, they design a residual reconstruction loss to ensure the complementary outputs from the two sub networks when training the model. 
Kim \textit{et al.} \cite{kimSingleImageReflection2020} first propose to use displacement mapping and path tracing to synthesize dataset with physically-based rendering. And they design a two-stage network which first separates the blending image to $T*$ and $\tilde{R}*$, then improves $\tilde{R}*$ by removing the glass/lens-effect. 
Specifically, Chang \textit{et al.} \cite{changSingleImageReflection2021} introduce a three-stage network with three auxiliary extensions: Edge Guidance, Reflection Classifier, and Recurrent Decomposition. They first obtain supplementary edge information, which provides more information to distinguish the two layers. They then train a reflection classifier to provide constraints and objectives for benefiting the model learning. Finally, a recurrent decomposition is proposed instead of adding more sub-networks.


\subsection{Vision Transformer}
Recently, natural language processing (NLP) model Transformer proposed by Vaswani \textit{et al.} has obtained superior performance against state-of-the-art methods in the computer vision community for various vision tasks. 
Transformer models have been successfully utilized for image recognition \cite{huLocalRelationNetworks2019,zhaoExploringSelfattentionImage2020,dosovitskiyImageWorth16x162020,touvronTrainingDataefficientImage2021}, object detection \cite{carionEndtoendObjectDetection2020,zhuDeformableDetrDeformable2020,liuDeepLearningGeneric2020,liuSwinTransformerHierarchical2021,liuSwinNetSwinTransformer2022,daiAO2DETRArbitraryOrientedObject2023}, image classification \cite{dosovitskiyImageWorth16x162020,liLocalvitBringingLocality2021,liuTransformerConvolutionalNeural2021,liuSwinTransformerHierarchical2021,vaswaniScalingLocalSelfattention2021,wuVisualTransformersTokenbased2020}, image segmentation \cite{yeCrossmodalSelfattentionNetwork2019,liuSwinTransformerHierarchical2021,wuVisualTransformersTokenbased2020,caoSwinunetUnetlikePure2021,zhengRethinkingSemanticSegmentation2021} and face restoration \cite{zhangBlindFaceRestoration2022,wang2022survey}.

Vision Transformer (ViT) \cite{dosovitskiyImageWorth16x162020} proposed by Dosovitskiy \textit{et al.} facilitates the transformation of backbone from CNNs to Transformers. This pioneering work has led to follow-up research aimed at improving its utility, but there are also limitations.
ViT is computationally expensive -- when encountering large-scale images, the time complexity required for its training is quadratic proportional to the image size. Wang \textit{et al.} \cite {wangPyramidVisionTransformer2021} propose a pyramid vision transformer (PVT) which utilizes a pyramid structure to extract multi-scale features for dense prediction tasks. Liu \textit{et al.} \cite{liuSwinTransformerHierarchical2021} propose a swin transformer which uses hierarchical feature maps similar to CNNs to obtain multi-scale features. They also introduce Windows Multi-Head Self-Attention (W-MSA) to calculate self-attention within each window, and Shifted Windows Multi-Head Self-Attention (SW-MSA), so that the feature extracted in every window can be transferred to adjacent windows. 
ViT is data costly -- it needs to be trained with a large amount of data to achieve its best performance. 
\cite{touvronTrainingDataefficientImage2021} proposes a teacher-student training strategy and token-based distillation. As a result, the proposed DeiT can achieve great results using the smaller-scale ImageNet-1K dataset for training.
In order to overcome the difficulty, Yuan \textit{et al.} \cite{yuanIncorporatingConvolutionDesigns2021} propose a convolution-enhanced image transformer (CeiT) which combines the advantage of CNNs in extracting low-level features, strengthening locality, and the advantages of Transformers in establishing long-range dependencies. 

Due to the success of Transformer-based models, there are also several transformer methods \cite{caoVideoSuperresolutionTransformer2021,chenPretrainedImageProcessing2021,wangUformerGeneralUshaped2022,zamirRestormerEfficientTransformer2022,zhang2022deep,wang2023ultra,wang2023gridformer} for image restoration. 
Chen \textit{et al.} \cite{chenPretrainedImageProcessing2021} propose an image processing transformer (IPT), which is a pre-trained model and achieves excellent performance on various low-level tasks. 
Cao \textit{et al.} \cite{caoVideoSuperresolutionTransformer2021} propose VSR-Transformer that utilizes a self-attention mechanism to restore high-resolution videos. 
Wang \textit{et al.} \cite{wangUformerGeneralUshaped2022} present a U-shaped architecture for image restoration which performs non-overlapping window-based self-attention instead of global self-attention. 
At the same time, Liang \textit{et al.} \cite{liangSwinirImageRestoration2021} introduce a strong baseline model called SwinIR for image restoration, which is based on the Swin Transformer \cite{liuSwinTransformerHierarchical2021}. 
In order to achieve stronger performance on more tasks, Restormer \textit{et al.} \cite{zamirRestormerEfficientTransformer2022} propose an efficient Transformer model by making several key designs in the building blocks which can learn long-range dependencies while remaining computationally efficient.


\section{Benchmark Datasets}
To benchmark current state-of-the-art SIRR methods, we propose two UHD datasets. In this section, we describe the features of our UHDRR datasets, and the generating method to construct these datasets. 

\subsection{The UHDRR Datasets}
These UHD images with 4K and 8K resolution are from \cite{zhangBenchmarkingUltraHighDefinitionImage2021} and cover various scenes. They are captured in indoor and outdoor scenarios by different cameras. Sample images in the UHDRR dataset are shown in Figure \ref{fig_dataset}. 
To ensure quality of the datasets, we have carefully checked all images and removed images with blurry background or poor lighting conditions. 

The resolution of images from UHDRR4K dataset is $3840 \times 2160$. 
It contains $2,999$ and $168$ image quadruplets for training and testing respectively. Specifically, a quadruplet is defined as ${T, R^*, R, B}$, where ${T}$ is the transmission image, ${R^*}$ is the reflection image, ${R}$ is the reflection mask image processed by random Gaussian smoothing kernel and ${B}$ is the blended image. 

The UHDRR8K dataset contains $1,014$ training image quadruplets and $105$ testing image quadruplets respectively. The resolution of image from UHDRR8K is $7680 \times 4320$. Similar to the UHDRR4K dataset, each of quadruplet consists of ${T, R^*, R, B}$.

\subsection{Image Synthesis Setting}
The synthesis method in this paper is the same as Zhang \textit{et al.} \cite{zhangSingleImageReflection2018}. 
We randomly divide the training set and testing set from \cite{zhangBenchmarkingUltraHighDefinitionImage2021} into two parts. One is used as the transmission layer ${T}$ and the other is set as the reflection layer ${R^*}$. 
From the principle of camera imaging, most of the reflection layer is usually out of focus because of the reflection of the glass, which causes the reflection layer to be more blurry and smoother than the transmission layer. 
Therefore, we apply a Gaussian smoothing kernel with random kernel size on the reflection layer ${R^*}$ to simulate this defocused reflection, which can be represented as 
\begin{equation}
    R = H_{Gaussian} \left ( R^* \right )
    ,\label{Gaussian}
\end{equation}
where ${R^*, R}$ are reflection layer and reflection mask layer, ${H_{Gaussian}}$ is the operation of random Gaussian smoothing.  
By doing so, we obtain the reflection mask layer ${R}$, which is used for subsequent synthesis operation. 
And when synthesizing transmission layer ${T}$ and reflection mask layer ${R}$ into blended layer ${B}$, we choose a random constant $\alpha$, representing the contribution of the transmission layer. 
It can be represented by the formula as 
\begin{equation}
	B = \left ( 1-\alpha \right ) \times R + \alpha \times T
	, \label{eq_blend}
\end{equation}
where ${T, R, B}$ are transmission layer, reflection mask layer, and blended layer, respectively. 
In our dataset, this random constant $\alpha$ is also provided in each quadruplet.

\section{The Proposed RRFormer Model}
\begin{figure*}[t]
	\centering
	\includegraphics[width=7 in]{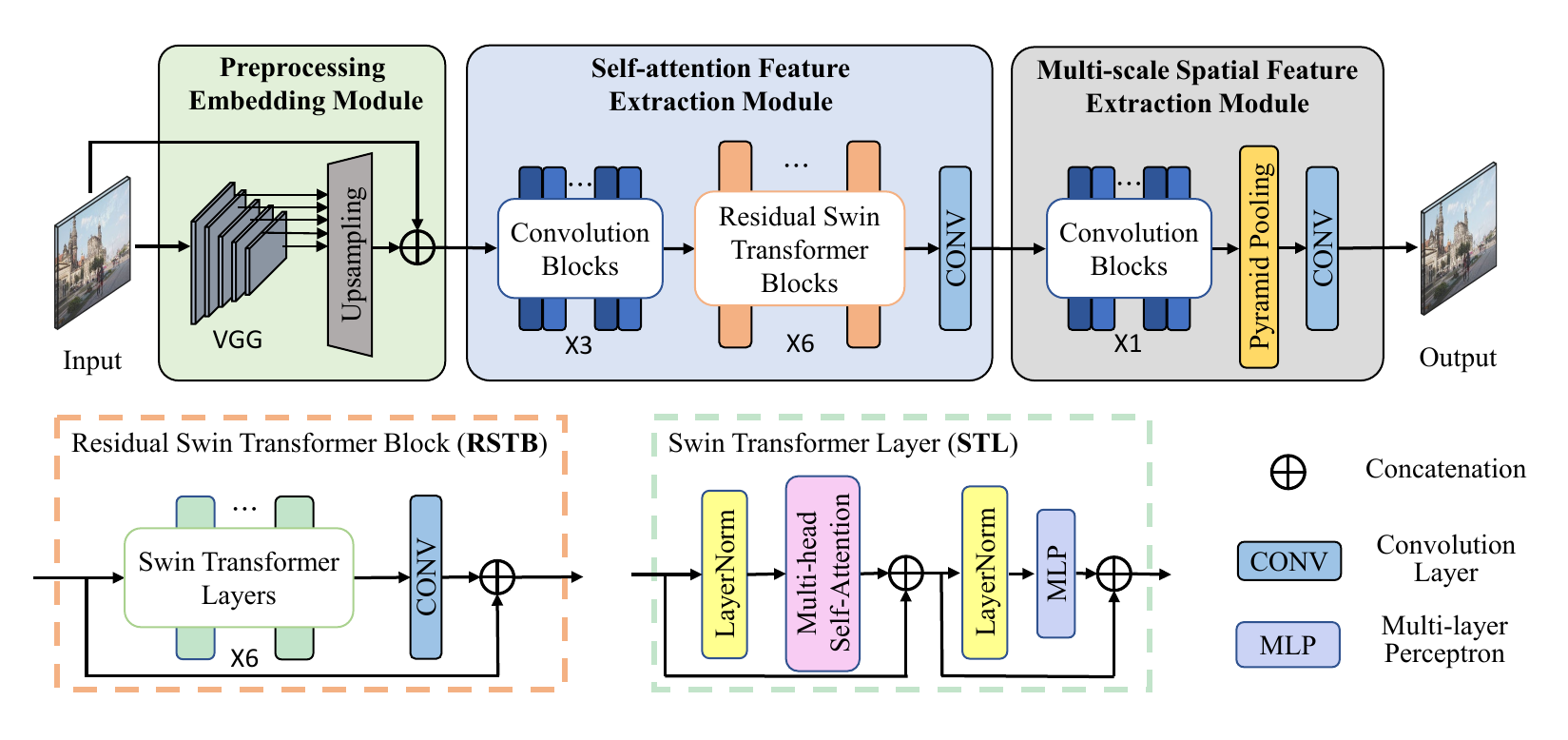}
	\caption{The architecture of the proposed RRFormer. It consists of three parts: Prepossessing Embedding Module, Self-attention Feature Extraction Module, and Multi-scale Spatial Feature Extraction Module. }
	\label{fig_network}
\end{figure*}

In this section, we introduce a transformer-based architecture for reflection removal named RRFormer. 
As shown in Figure \ref{fig_network}, the proposed RRFormer consists of three parts: the Prepossessing Embedding Module, the Self-attention Feature Extraction Module, and the Multi-scale Spatial Feature Extraction Module.

\subsection{Network Architecture}

\textbf{Preprocessing Embedding Module}. 
Given an input blended image $I^\star\in \mathbb{R}^{ H\times W\times C_{I}}$ ($H$ is the height of the image, $W$ is the width of the image, and $C_{I}$ is the channel number), the network first applies a pretrained VGG-19 network \cite{szeliskiLayerExtractionMultiple2000} to extract hypercolumn \cite{hariharanHypercolumnsObjectSegmentation2015} features as 
\begin{equation}
	F_{C0}=H_{C0} \left ( I^\star \right ) 
	, \label{eq1}
\end{equation}
where $F_{C0}\in \mathbb{R}^{ H\times W\times C}$ is the hypercolumn features and $C$ is the channel number of this feature. 
$H_{C0}$ is a function indicating the first convolutional layer.
Then we concatenate the input blended image with these hypercolumn features $F_{C0}$ as an enhanced network input, which can be represented as 
\begin{equation}
	F_{C1} = F_{C0}\bigoplus I^\star 
	. \label{eq2}
\end{equation}
$\bigoplus$ and $F_{C1}$ are the concatenation operation, and the enhanced network input, respectively. 

\textbf{Self-attention Feature Extraction Module}. 
The enhanced network input is forwarded to a feature extraction module to obtain deep features. The feature extraction module comprises several convolutional layers, ReLU layers and residual Swin Transformer blocks consisting of several Swin Transformer layers followed by a convolutional layer. 
In this module, we first forward $F_{C1}$ into three consecutive convolutional layers and ReLU layers, \textit{i.e}. $C1, R1, C2, R2, C3, R3$, which are expressed by the equation as
\begin{equation}
	F_{C2}=      H_{R1}\left (  H_{C1} \left ( F_{C1} \right )  \right )  
	, \label{eq8}
\end{equation}
\begin{equation}
	F_{C3}=H_{R2}\left (  H_{C2} \left ( F_{C2} \right )  \right )  
	, \label{eq9}
\end{equation}
\begin{equation}
	F_{C4}=H_{R3}\left (  H_{C3} \left ( F_{C3} \right )  \right )  
	, \label{eq10}
\end{equation}
where $H_{C1}, H_{C2}, H_{C3}$ are three convolutional layers, $H_{R1}, H_{R2}, H_{R3}$ are three ReLU layers, and $F_{C2}, F_{C3}, F_{C4}$ are the output of each of the above equations. 
Then several residual Swin Transformer blocks are applied to extract global features as 
\begin{equation}
	F_{RSTB}=H_{RSTB}\left (F_{C4}\right )  
	, \label{eq_RSTB}
\end{equation}
where $H_{RSTB}$ is the operation of residual Swin Transformer blocks and $F_{RSTB}$ is the corresponding feature. The details of $H_{RSTB}$ are introduced in the next subsection \ref{RSTB}. 

\textbf{Multi-scale Spatial Feature Extraction Module. }
Although Swin transformer Layer (STL) \ref{STL} can utilize regular and shifted window partitioning alternately to facilitate cross-window connections \cite{liuSwinTransformerHierarchical2021}, leveraging complementary multi-scale spatial information can yield further advantages. 
To accomplish this, we utilize a pyramid pooling module \cite{heSpatialPyramidPooling2015,zhaoPyramidSceneParsing2017} to aggregate contextual information from different regions, thus improving the ability to obtain global information. 
In this study, we reset the pyramid pooling scale with bin sizes of 4, 8, 16, and 32, respectively, as illustrated in Figure \ref{fig_network}. The output feature maps of different levels within the pyramid pooling module have varying sizes. Therefore, we upsample the low-dimensional feature maps to acquire the same-sized feature as the original feature map. Subsequently, the output of the pyramid pooling module comprises four distinct pyramid scales of features concatenated together. 

In addition, we need a non-linear transformation (\textit{i.e.}, a Conv-ReLU pair) before the pyramid pooling module to adjust the channel dimension, which can be formulated as
\begin{equation}
	F_{M1}=H_{R_{M1}}\left (  H_{C_{M1}} \left ( F_{RSTB} \right ) \right ) 
	, \label{eq_convrelu}
\end{equation}
\begin{equation}
	F_{pyramid}=H_{pyramid} \left ( F_{M1} \right )  
	, \label{eq_pyramid}
\end{equation}
where $H_{C_{M1}}, H_{R_{M1}}$ are convolutional layer and ReLU layer, respectively. $F_{M1}, F_{pyramid}$ is the dimension-adjusted features and multi-scale spatial features, respectively. $H_{pyramid}$ denotes the operation of pyramid pooling. 
Then a convolutional layer is applied to $F_{pyramid}$ to reconstruct transmission layer images. The process can be described as
\begin{equation}
	F_{T_{f}}=H_{C_{down}} \left ( F_{pyramid} \right )  
	, \label{eq_RRFormer}
\end{equation}
where $H_{C_{down}}$ is the operation to reduce the channel dimension and $F_{T_{f}}$ denotes the predicted transmission layer image. The final output of RRFormer is $F_{T_{f}}$.

\subsection{Residual Swin Transformer Block}\label{RSTB}
Following \cite{liangSwinirImageRestoration2021}, we apply the residual swin transformer block in our RRFormer to extract different levels of features. Given the input feature $F_{0}$, we first feed it to $N$ Swin Transform layers as 
\begin{equation}
	F_{i} = H_{{STL}_{i}}\left ( L_{i-1} \right ) , i = 1,2,\cdots,N, \label{eq3}
\end{equation}
where $H_{STL_{i}\left ( \cdot \right )}$ is the $i$-th swin transformer layer, $L_{i}$ and $L_{i-1}$ are its input and output. Therefore, $F_{N}$ is the output of $N$ Swin Transform layers. 
Finally, we apply a convolutional layer before the residual connection, which can be formulated as
\begin{equation}
	F_{output} = H_{CONV}\left ( L_{N} \right ) + F_{0} , \label{eq4}
\end{equation}
where $F_{0}$ and $H_{CONV}\left ( \cdot \right )$ are the input feature and convolution operation, respectively.

\begin{figure*}[t]
	\centering
	\subfloat
	{
		\includegraphics[width=7in]{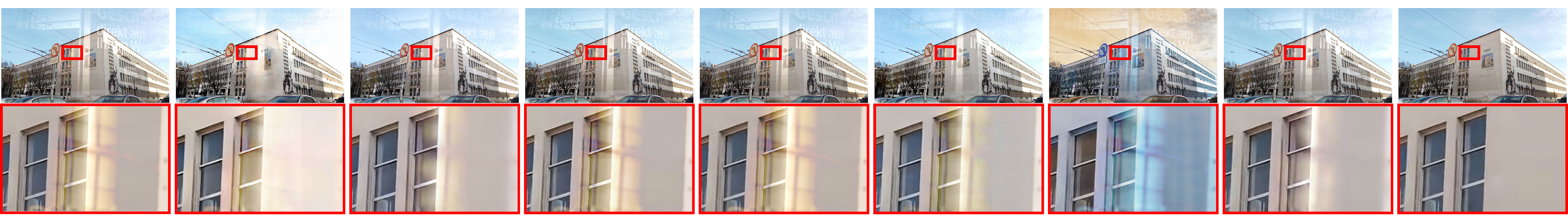}
		\label{fig_4k_1}
	}
	\vspace{-3mm}
	\subfloat
	{
		\includegraphics[width=7in]{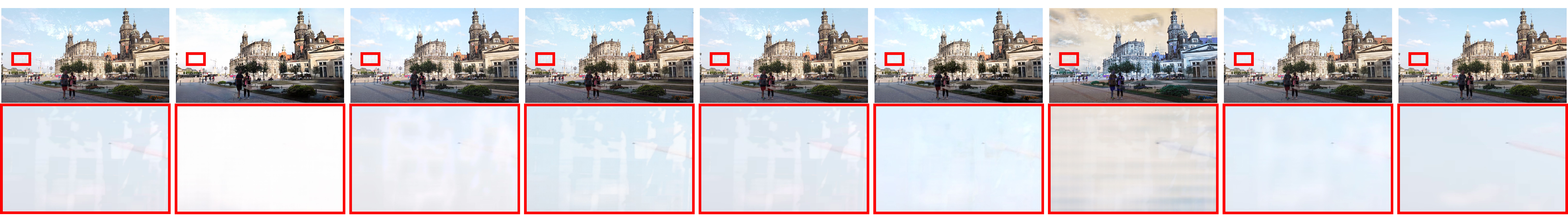}
		\label{fig_4k_2}
	}
	\vspace{-3mm}
	\subfloat
	{
		\includegraphics[width=7in]{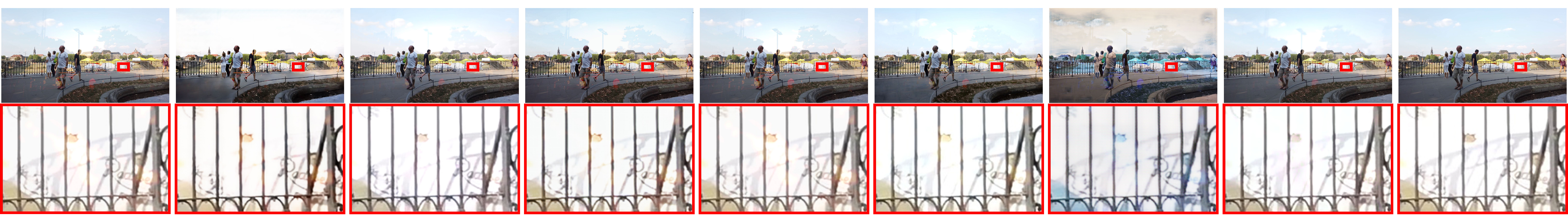}
		\label{fig_4k_3}
	}
	\vspace{-3mm}
	\subfloat
	{
		\includegraphics[width=7in]{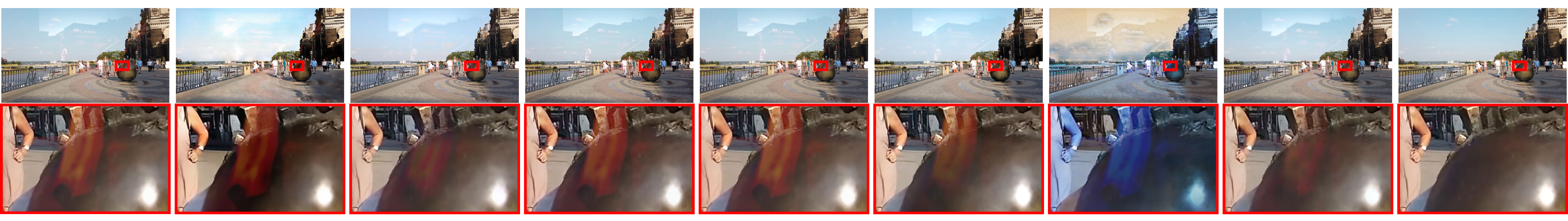}
		\label{fig_4k_4}
	}
	\vspace{-3mm}
	\subfloat
	{
		\includegraphics[width=7in]{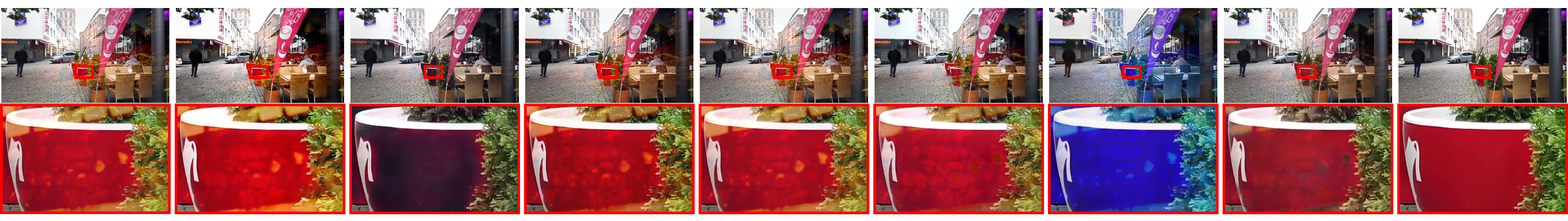}
		\label{fig_4k_5}
	}
	\vspace{-3mm}
	\subfloat
	{
		\includegraphics[width=7in]{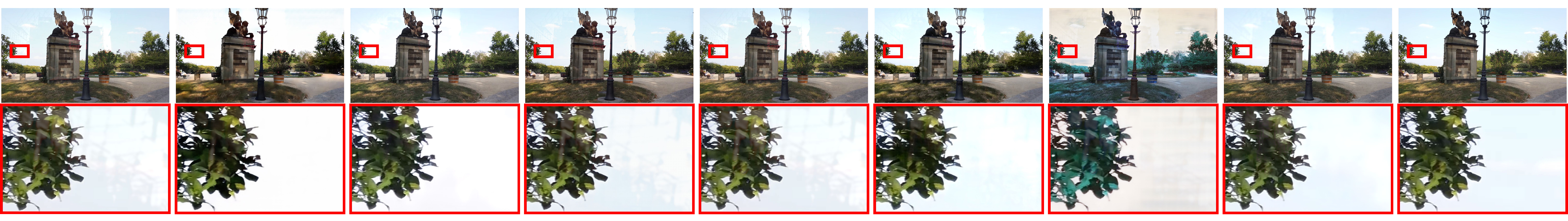}
		\label{fig_4k_6}
	}
	\vspace{-3mm}
	\subfloat
	{
		\includegraphics[width=7in]{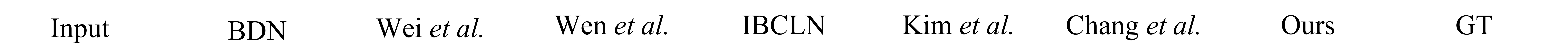}
		\label{fig_4k_name}
	}
	\caption{Visual results on the UHDRR4K dataset. From left to right are the input, the results of BDN \cite{yangSeeingDeeplyBidirectionally2018}, Wei \textit{et al.} \cite{weiSingleImageReflection2019}, Wen \textit{et al.} \cite{wenSingleImageReflection2019}, IBCLN \cite{liSingleImageReflection2020}, Kim \textit{et al.} \cite{kimSingleImageReflection2020}, Chang \textit{et al.} \cite{changSingleImageReflection2021}, ours, and ground-truth. Best viewed in color. }
	\label{fig_4k}
\end{figure*}
\begin{figure*}[t]
	\centering
	\subfloat
	{
		\includegraphics[width=7in]{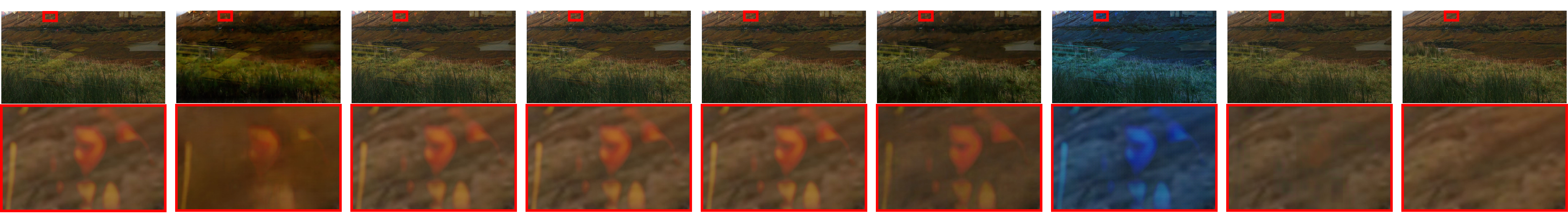}
		\label{fig_8k_1_1}
	}
	\vspace{-3mm}
	\subfloat
	{
		\includegraphics[width=7in]{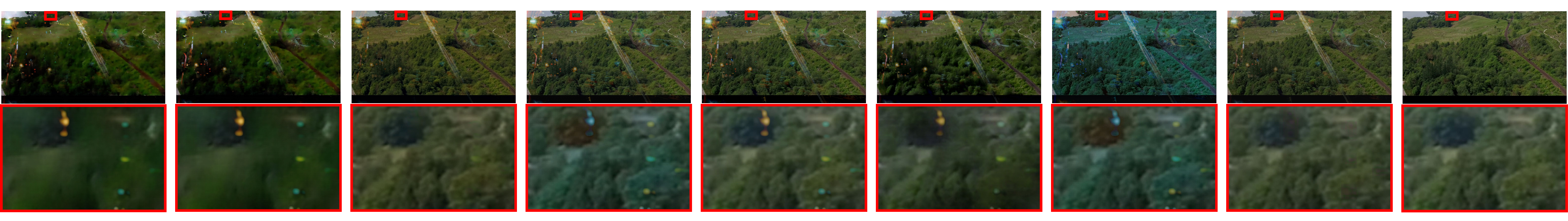}
		\label{fig_8k_1_2}
	}
	\vspace{-3mm}
	\subfloat
	{
		\includegraphics[width=7in]{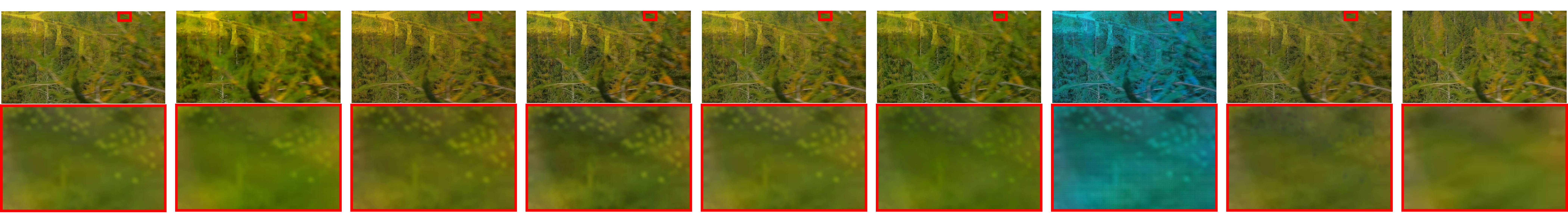}
		\label{fig_8k_1_3}
	}
	\vspace{-3mm}
	\subfloat
	{
		\includegraphics[width=7in]{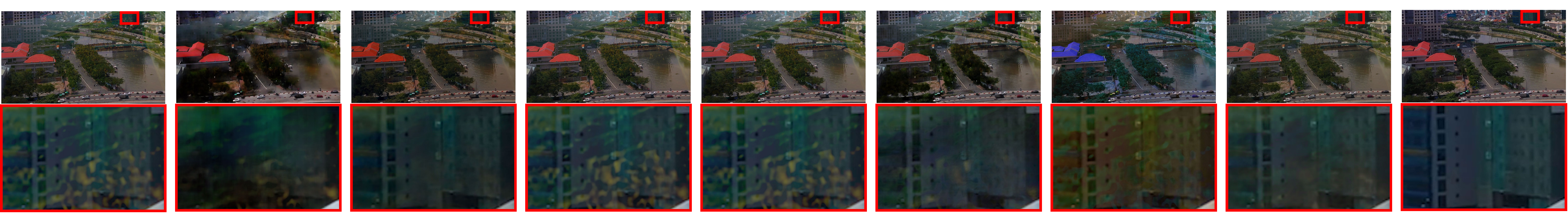}
		\label{fig_8k_1_4}
	}
	\vspace{-3mm}
	\subfloat
	{
		\includegraphics[width=7in]{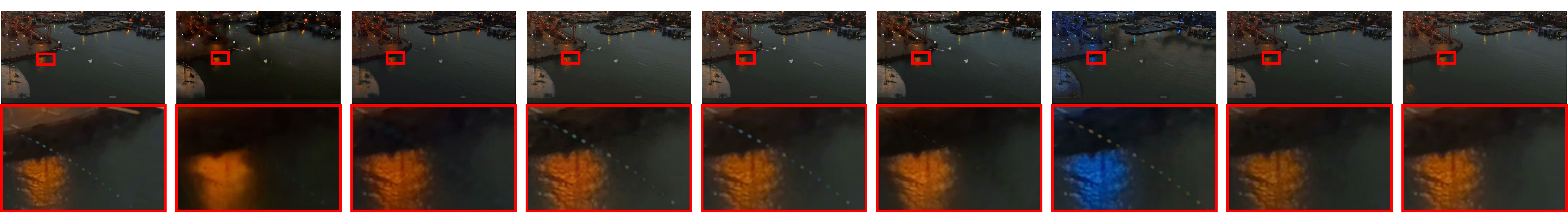}
		\label{fig_8k_1_5}
	}
	\vspace{-3mm}
	\subfloat
	{
		\includegraphics[width=7in]{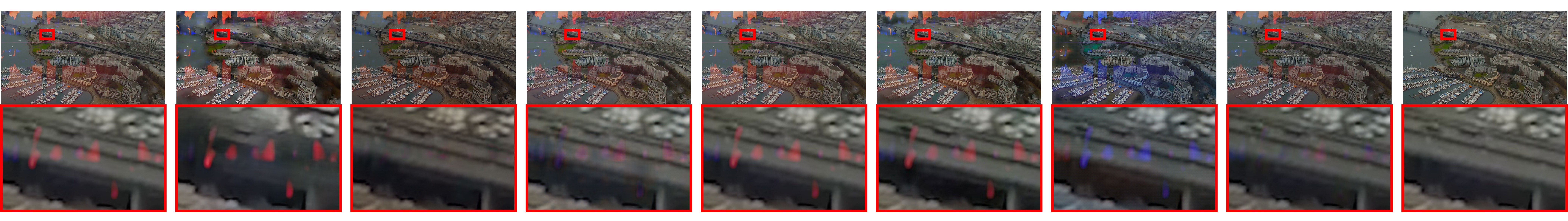}
		\label{fig_8k_1_6}
	}
	\vspace{-3mm}
	\subfloat
	{
		\includegraphics[width=7in]{4k-1-result/4k-1-name.pdf}
		\label{fig_8k_name}
	}
	\caption{Visual results on the UHDRR8K dataset. From left to right are the input, the results of BDN \cite{yangSeeingDeeplyBidirectionally2018}, Wei \textit{et al.} \cite{weiSingleImageReflection2019}, Wen \textit{et al.} \cite{wenSingleImageReflection2019}, IBCLN \cite{liSingleImageReflection2020}, Kim \textit{et al.} \cite{kimSingleImageReflection2020}, Chang \textit{et al.} \cite{changSingleImageReflection2021}, ours, and ground-truth. Best viewed in color. }
	\label{fig_8k}
\end{figure*}
\begin{table}[t]
	\centering
	\caption{Performance comparison of representative methods for SIRR on the UHDRR4K dataset. Both PSNR and SSIM values are reported. }
	\label{table_4k}
	\begin{tabular}{l|rrr}
		\toprule
		&PSNR&SSIM\\
		\midrule
		BDN \cite{yangSeeingDeeplyBidirectionally2018}&19.81&0.897\\
		Wei \textit{et al.} \cite{weiSingleImageReflection2019}&\underline{\textit{24.37}}&\textbf{0.974}\\
		Wen \textit{et al.} \cite{wenSingleImageReflection2019}&19.53&0.928\\
		IBCLN \cite{liSingleImageReflection2020}&24.12&0.968\\
		Kim \textit{et al.} \cite{kimSingleImageReflection2020}&23.45&0.929\\
		Chang \textit{et al.} \cite{changSingleImageReflection2021}&22.93&0.914\\
		\midrule
		Ours&\textbf{24.71}&\underline{\textit{0.971}}\\
		\bottomrule
	\end{tabular}
\end{table}

\textbf{Swin Transformer Layer (STL).} \label{STL}
\cite{liuSwinTransformerHierarchical2021} introduces the swin transformer blocks, which is an improvement on the original transformer layer \cite{vaswaniAttentionAllYou2017}. It applies the shifted window mechanism to obtain local attention rather global attention. Given an input $F_{0}\in \mathbb{R}^{ H\times W\times C}$, the Swin block divides it into $M \times M$ local windows. Then it obtains $HM/M^{2}$ features whose size is $M^{2} \times C$. For a local feature, it computes similarity as
\begin{equation}
	\operatorname{Attention}(Q,K,V)=\operatorname{SoftMax}(QK^{T}/\sqrt{d}+B)V,  \label{eq5}
\end{equation}
where $Q, K, V\in \mathbb{R}^{M^{2}\times d}$ are the \textit{query, key} and \textit{value} matrices; $B$ is the learnable relative positional encoding and $d$ is the \textit{query/key} dimension. It also applies shifted window multi-head self-attention (MSA) to overcome the problem that information cannot be passed between windows. 

Then, a two-layer multi-layer perception (MLP) with ReLU non-linearity in between is utilized for further extracting features. A LayerNorm (LN) layer is added before both MSA and MLP, and a residual connection is applied for both modules. The whole process is illustrated as 
\begin{equation}
	F_{MSA} = \operatorname{MSA}(\operatorname{LN}(F_{local})) + F_{local},  \label{eq6}
\end{equation}
\begin{equation}
	F_{MLP} = \operatorname{MLP}(\operatorname{LN}(F_{MSA})) + F_{MSA},  \label{eq7}
\end{equation}
where $\operatorname{MLP}, \operatorname{LN}$ and $MLP$ are the functions indicating multi-head self-attention, LayerNorm, and multi-layer perception; $F_{local}, F_{MSA}$ and $F_{MLP}$ are the input of local feature, the output of MSA, and the output of MLP, respectively.

\subsection{Loss Function}
Following \cite{weiSingleImageReflection2019}, our loss function contains three terms: pixel loss, feature loss, and adversarial loss.

\textbf{Pixel loss.} 
To minimize the difference between $T$ and $T^\star$, we first apply the mean squared errors (MSE). 
Then, we also consider the discrepancy of gradients between $T$ and $T^\star$. Let the symbol $T^\star$ denote the ground truth, the loss function can be represented as 
\begin{equation}
\begin{aligned}
	\mathcal{L}_{pixel} &= \alpha \left \| T - T^\star \right \| ^2_{2} \\&+\beta \left (  \left \| \nabla _{x}T - \nabla _{x}T^\star  \right \|_{1} + \left \| \nabla _{y}T - \nabla _{y}T^\star  \right \|_{1} \right ) ,
\end{aligned}  
	\label{Lpixel}
\end{equation}
where $\alpha$ and $\beta$ are constants, $\nabla _{x}$ and $\nabla_{y}$ denote the gradient operator. The gradient discrepancy is applied to reduce blurry images \cite{narihiraDirectIntrinsicsLearning2015}. 
For all experiments, we set $\alpha = 0.2$ and $\beta = 0.4$. 

\textbf{Feature loss. }
To measure the difference between $T$ and $T^\star$, we utilize the pre-trained VGG-19 network $\Phi$ to obtain the low-level and high-level features. The feature loss is defined as 
\begin{equation}
    \mathcal{L}_{feat} = \sum_{l}\lambda_{l}\left \| \Phi_{l}\left ( T \right ) - \Phi_{l} \left ( T^\star \right ) \right \| _{1},
   	\label{Lfeat}
\end{equation}
where $\Phi_{l}$ is the layer $l$ in the pre-trained VGG-19 network and $\lambda_{l}$ indicates the balancing weight. 
Similar to \cite{zhangSingleImageReflection2018}, we select the layers ``conv2\_2", ``conv3\_2", ``conv4\_2" and ``conv5\_2" in the VGG-19 network. 

\textbf{Adversarial loss. }
To prevent the network from generating unreal images, an adversarial loss is necessary. We utilize a relativistic discriminator \cite{jolicoeur-martineauRelativisticDiscriminatorKey2018} which uses both real data and fake data to measure the probability from absolute truth to relative truth. We define the adversarial loss as
\begin{equation}
\begin{aligned}
	\mathcal{L}_{adv} =  &-\log\left ( SIGMOD\left ( C\left ( T \right ) -C\left ( T^\star \right )  \right )  \right ) \\ 
 &-\log\left ( 1- SIGMOD\left ( C\left ( T^\star \right ) -C\left ( T \right ) \right ) \right ) ,
 \end{aligned}
	 \label{Ladv}
\end{equation}
where $C\left ( \cdot  \right )$ is the non-transormed discriminator function.

In summary, we empirically set the coefficients of each loss term, and then the final loss function is given as follows, 
\begin{equation}
	\mathcal{L}_{all} = \mathcal{L}_{pixel} + \lambda_{1}\mathcal{L}_{feat} + \lambda_{2}\mathcal{L}_{adv},  \label{eq11}
\end{equation}
where the weights $\lambda_{1}$ and $\lambda_{2}$ are set as $0.1$ and $0.01$ respectively in all experiments.

\section{Experiments and Analysis}
In this section, we first benchmark existing SIRR methods and our proposed RRFormer on the UHDRR4K and UHDRR8K datasets, and then evaluate the proposed RRFormer on the CDR dataset \cite{leiCategorizedReflectionRemoval2022}. 

\subsection{Evaluated SIRR Methods}\label{Evaluated_Methods}
In this benchmark study, we evaluate six state-of-the-art SIRR methods, BDN \cite{yangSeeingDeeplyBidirectionally2018}, Wei \textit{et al.} \cite{weiSingleImageReflection2019}, Wen \textit{et al.} \cite{wenSingleImageReflection2019}, IBCLN \cite{liSingleImageReflection2020}, Kim \textit{et al.} \cite{kimSingleImageReflection2020}, and Chang \textit{et al.} \cite{changSingleImageReflection2021}.

These methods are of diverse network structures and they have achieved great performance on several datasets, \textit{i.e.}, SIR$^2$ \cite{wanBenchmarkingSingleimageReflection2017}, Zhang \textit{et al.} \cite{zhangSingleImageReflection2018}, Nature \cite{liSingleImageReflection2020}, and CDR \cite{leiCategorizedReflectionRemoval2022}.  BDN \cite{yangSeeingDeeplyBidirectionally2018} and Kim \textit{et al.} \cite{kimSingleImageReflection2020} are two representative cascade neural network architectures to predict both transmission and reflection layers of input images. 
Wen \textit{et al.} \cite{wenSingleImageReflection2019} additionally predicts alpha blending masks. 
Wei \textit{et al.} \cite{weiSingleImageReflection2019} and IBCLN \cite{liSingleImageReflection2020} are typical physically-based methods, considering the alignment-invariant and the spatial variability respectively. 
Chang \textit{et al.} \cite{changSingleImageReflection2021} is a novel decomposition model, decomposing the blended image into transmission layer and reflection layer. 

All these deep SIRR methods are re-trained on our proposed datasets, except BDN \cite{yangSeeingDeeplyBidirectionally2018} and Kim \textit{et al.} \cite{kimSingleImageReflection2020}. As these two works do not provide training code, we use their pre-trained models to evaluate on the UHDRR datasets. 

\subsection{Implementation}
In this paper, all the methods are trained for $100$ epochs using V100 GPU. We set the learning rate to $0.0002$ for all methods. When training the networks on our dataset, to make it more comprehensive, we follow the practice in \cite{weiSingleImageReflection2019} to add $90$ real-world images from \cite{zhangSingleImageReflection2018}. 
Patches of size $256 \times 256$ are randomly cropped from the images in the fusion dataset.
In the testing stage, we take the whole 4K image as input. Additionally, 8K images are cropped to four non-overlap 4K-resolution patches. In this paper, we use SSIM and PSNR as quantitative metrics to assess the quantity between the predicted transmission layer and the corresponding groundtruth.

\subsection{Results on UHDRR4K Dataset}
We first evaluate the current SOTA methods mentioned in Section \ref{Evaluated_Methods} and our proposed RRFormer on the UHDRR4K dataset to investigate their performance for the task of 4K image reflection removal. Table \ref{table_4k} shows the quantitative comparison in terms of PSNR and SSIM.
The first and second best results are marked by bold font and italic font with underline, respectively. 
Among the seven methods of SIRR, our proposed RRFormer achieves the best performance regarding the PSNR metric, with an advantage of $0.34$db over the second best method.
In terms of SSIM, RRFormer obtains the second best performance, with a decrease of merely $0.003$ compared to the best performance \cite{weiSingleImageReflection2019}. 
IBCLN \cite{liSingleImageReflection2020} achieves the third best performance in terms of both PSNR and SSIM. 

We also show a visual comparison among different methods on the UHDRR4K dataset in Figure \ref{fig_4k}. In general, our proposed RRFormer generates images with finer details. 
Though RRFormer is inferior to the best method by Wei \textit{et al.} \cite{weiSingleImageReflection2019} in terms of SSIM, it produces images with better visual quality. For example, in the fifth example of Figure \ref{fig_4k}, the method by Wei \textit{et al.} \cite{weiSingleImageReflection2019} fails to recover the true tonality, while our RRFormer succeeds.

It is notable that, in terms of tonal performance, the result images recovered by Chang \textit{et al.} \cite{changSingleImageReflection2021} are significantly different from the groundtruth. The result images by BDN \cite{yangSeeingDeeplyBidirectionally2018} are slightly bright and thus cause the lost of details. 
RRFormer performs better when the reflection appears in the sky region. Compared with these two methods, RRFormer produces more real images with better tone fidelity.
\begin{table*}[t]
	\caption{Performance comparison of representative methods for SIRR on the CDR dataset, including the 'All' dataset and all other sub-datasets. 
	Both PSNR and SSIM values are reported. The first and second best results are marked by bold font and italic font with underline, respectively. \label{table_cdr}}
	\centering
	\begin{tabular}{lcccccccc}
		\toprule
		&\multicolumn{2}{c}{ALL}&\multicolumn{2}{c}{SRST}&\multicolumn{2}{c}{BRST}&\multicolumn{2}{c}{Non-ghosting}\\
		&PSNR&SSIM&PSNR&SSIM&PSNR&SSIM&PSNR&SSIM\\
		\midrule
		Li \textit{et al.} \cite{liSingleImageLayer2014}&12.73& 0.650& 12.26 & 0.565& 13.19& 0.723& 12.56& 0.624\\
		Arvan. \textit{et al.} \cite{arvanitopoulosSingleImageReflection2017}&19.63& 0.753& 18.24 &0.680 & 20.91& 0.816& 19.00& 0.727\\
		Yang \textit{et al.} \cite{yangFastSingleImage2019}&19.42& 0.767 & 18.10 & 0.680 & 20.65& 0.841& 18.78& 0.738 \\
		\midrule
		CEILNet \cite{fanGenericDeepArchitecture2017}&17.96& 0.708& 16.17 & 0.596& 19.49& 0.802& 17.24& 0.673\\
		Zhang \textit{et al.} \cite{zhangSingleImageReflection2018}&15.20& 0.694& 13.52 & 0.590& 16.58& 0.780& 14.48& 0.662\\
		BDN \cite{yangSeeingDeeplyBidirectionally2018}&18.97& 0.758& 19.04 & \underline{\textit{0.713}}& 19.06& 0.799& 18.62& 0.733\\
		Wei \textit{et al.} \cite{weiSingleImageReflection2019}&\underline{\textit{21.01}} & 0.762& 19.52& 0.672& 22.36 & 0.839& \underline{\textit{20.50}} & 0.731\\
		CoRRN \cite{wanCoRRNCooperativeReflection2019}&20.22& \underline{\textit{0.774}} & \textbf{20.32}& 0.699 & 20.08& 0.838& 20.37& \underline{\textit{0.750}} \\
		IBCLN \cite{liSingleImageReflection2020}&19.85& 0.764& 18.33 & 0.671& 21.14&\underline{\textit{0.842}} & 19.23& 0.735\\
		Kim \textit{et al.} \cite{kimSingleImageReflection2020}&21.00 & 0.760&19.27& 0.676&\underline{\textit{22.61}}&0.833&20.42 & 0.731\\
		\midrule
		Ours&\textbf{22.33}& \textbf{0.815}& \underline{\textit{19.59}} & \textbf{0.739}& \textbf{23.04}& \textbf{0.866} & \textbf{21.25}& \textbf{0.791}\\
		\bottomrule
		\\ \hspace*{\fill} \\
		\toprule
		&\multicolumn{2}{c}{Weak $R$}&\multicolumn{2}{c}{Moderate $R$}&\multicolumn{2}{c}{Strong $R$}&\multicolumn{2}{c}{Ghosting}\\
		&PSNR&SSIM&PSNR&SSIM&PSNR&SSIM&PSNR&SSIM\\
		\midrule
		Li \textit{et al.} \cite{liSingleImageLayer2014}&14.36& 0.779& 12.47& 0.636& 8.89& 0.309& 13.36& 0.742\\
		Arvan. \textit{et al.} \cite{arvanitopoulosSingleImageReflection2017}&23.52& 0.878& 18.43& 0.744& 13.56& 0.397& 21.88& 0.844\\
		Yang \textit{et al.} \cite{yangFastSingleImage2019}&23.18& \underline{\textit{0.903}} & 18.28& 0.754& 13.50& 0.402& 21.72& \underline{\textit{0.870}} \\
		\midrule
		CEILNet \cite{fanGenericDeepArchitecture2017}&21.34& 0.862& 17.02& 0.685& 12.06& 0.341& 20.51& 0.836\\
		Zhang \textit{et al.} \cite{zhangSingleImageReflection2018}&17.20& 0.827& 15.10& 0.685& 9.33& 0.311& 17.81& 0.806\\
		BDN \cite{yangSeeingDeeplyBidirectionally2018}&21.10& 0.867& 18.25& 0.746& 16.15 & \underline{\textit{0.485}} & 20.20& 0.850\\
		Wei \textit{et al.} \cite{weiSingleImageReflection2019}&24.89 & 0.901& 19.42& 0.737& \underline{\textit{17.00}} & 0.450& \underline{\textit{22.80}}& \textbf{0.871} \\
		CoRRN \cite{wanCoRRNCooperativeReflection2019}&20.50& 0.890& \textbf{21.01} & \underline{\textit{0.768}} & 15.12& 0.433& 19.70& 0.861\\
		IBCLN \cite{liSingleImageReflection2020}&23.17& 0.899& 18.98& 0.752& 13.81& 0.395& 22.07& 0.867\\
		Kim \textit{et al.} \cite{kimSingleImageReflection2020}&\underline{\textit{25.03}} & 0.897& 19.66& 0.740& 15.25& 0.431& \textbf{23.10} & 0.865\\
		\midrule
		Ours&\textbf{25.50}& \textbf{0.912} & \underline{\textit{19.78}} & \textbf{0.782} & \textbf{20.09}& \textbf{0.507} & 21.35 & 0.851\\
		\bottomrule
	\end{tabular}
\end{table*}

\subsection{Results on UHDRR8K Dataset}
To benchmark the six methods of SIRR in the scenery of 8K images, we provide quantitative results on the UHDRR8K dataset in Table \ref{table_8k}. The first and second best results are marked by bold font and italic font with underline, respectively. In terms of PSNR, RRFormer achieves the best performance, with advance of $0.40$db compared to the second best method. Regarding SSIM, both RRFormer and Wei \textit{et al.} \cite{weiSingleImageReflection2019} obtain the best performance, slightly better than the second best one. 
Figure \ref{fig_8k} shows the visual comparison between different methods on the UHDRR8K dataset. Although the results of RRFormer do not show difference compared with Chang \textit{et al.} \cite{changSingleImageReflection2021} in terms of reflection removal, RRFormer produces more realistic images when generating the transmission layer and the color tone is closer to ground truth. 

In Figure \ref{fig_8k}, the results show that all the transmission layers predicted by Wei \textit{et al.} \cite{weiSingleImageReflection2019}, Wen \textit{et al.} \cite{wenSingleImageReflection2019} and Kim \textit{et al.} \cite{kimSingleImageReflection2020} exhibit obvious shadows and partial reflections, but IBCLN \cite{liSingleImageReflection2020}, Chang \textit{et al.} \cite{changSingleImageReflection2021} and RRFormer perform well.

\begin{table}[t]
	\centering
	\caption{Performance comparison of representative methods for SIRR on the UHDRR8K dataset. Both PSNR and SSIM values are reported. }
	\label{table_8k}
	\begin{tabular}{l|rrr}
		\toprule
		&PSNR&SSIM\\
		\midrule	
		BDN \cite{yangSeeingDeeplyBidirectionally2018}&19.41&0.851\\
		Wei \textit{et al.} \cite{weiSingleImageReflection2019}&\underline{\textit{22.01}}&\textbf{0.942}\\
		Wen \textit{et al.} \cite{wenSingleImageReflection2019}&16.79&0.866\\
		IBCLN \cite{liSingleImageReflection2020}&21.76&\underline{\textit{0.941}}\\
		Kim \textit{et al.} \cite{kimSingleImageReflection2020}&20.76&0.910\\
		Chang \textit{et al.} \cite{changSingleImageReflection2021}&21.31&0.899\\
		\midrule
		Ours&\textbf{22.41}&\textbf{0.942}\\
		\bottomrule
	\end{tabular}
\end{table}

\subsection{CDR Dataset}
To make our study more convincing, we also evaluate our proposed RRFormer on the public non-UHD CDR dataset \cite{leiCategorizedReflectionRemoval2022}. 
It provides an `ALL' dataset with $1,063$ triplets. And it also splits the `ALL' dataset into multiple sub-datasets according to smoothness, relative intensity, and the ghost, such as SRST (sharp reflection and sharp transmission), BRST (blurry reflection and sharp transmission), Non-ghosting, Weak $R$, Moderate $R$, Strong $R$ and Ghosting. 

We train and test on the `ALL' dataset and other sub-datasets separately. The results are shown in Table \ref{table_cdr}. 
For the `ALL' dataset, compared to existing methods, our proposed RRFormer outperforms all other methods in terms of both PSNR and SSIM, with the advance of $1.32$dB (PSNR) over Wei et al. \cite{weiSingleImageReflection2019} and $0.041$ (SSIM) over CoRRN \cite{wanCoRRNCooperativeReflection2019}.

For sub-datasets, RRFormer also performs better compared to other methods overall. 
Among these state-of-the-art methods, in terms of PSNR, CoRRN \cite{wanCoRRNCooperativeReflection2019} achieves the best performance in sub-datasets SRST and Moderate $R$. Kim \textit{et al.} achieve the best performance in Ghosting. Also, RRFormer outperforms other methods in BRST, Non-ghosting, Weak $R$ and Strong $R$. In terms of SSIM, Wei \textit{et al.} achieve the best performance in Ghosting, and RRFormer achieves the best performance in all other sub-datasets. Especially in Non-ghosting, RRFormer improves the SSIM by almost $0.04$ compared with the second best.

\section{Conclusion}
In this paper, we explore the domain of single image reflection removal in the scenery of UHD resolution. 
We present two new large-scale UHDRR datasets, UHDRR4K and UHDRR8K, which are the first UHD image datasets for benchmarking SIRR methods. Our dataset contains images of various scenes, which are captured indoor and outdoor by different cameras. Each of train/test samples is a quadruplet consisting of transmission	image, reflection image, reflection mask image, and blended image. To facilitate subsequent works, we also provide the blending ${\alpha}$ in each quadruplet. 
To explore the performance of  current SIRR methods on the UHD datasets, we evaluate six state-of-the-art SIRR methods.
We also propose a transformer-based architecture for reflection removal, named as RRFormer for SIRR task which performs satisfactorily on the CDR dataset and our UHDRR datasets. 
In the future, we will explore more advanced SIRR models for UHD resolutions. 

\section*{Acknowledgment}
This work was supported in part by the Jiangsu Funding Program for Excellent Postdoctoral Talent under Grant 2022ZB268.

\bibliography{books}

\end{document}